 \documentclass[final,3p,times,twocolumn]{elsarticle}

\usepackage{natbib}
\setcitestyle{authoryear,open={(},close={)},citesep={;}}

\usepackage{amssymb}
\usepackage{hyperref}
\usepackage{graphicx}
\usepackage{subcaption}
\usepackage{lscape}
\usepackage{amsmath}
\usepackage{multirow}
\usepackage[figuresright]{rotating}
\usepackage{lineno}
\usepackage{algorithm} 
\usepackage{algpseudocode} 
\usepackage{comment}
\usepackage{lineno}

\begin{document}

\begin{frontmatter}



\title{O2RNet: Occluder-Occludee Relational Network for Robust Apple Detection in Clustered Orchard Environments}


\author[label1]{Pengyu Chu}
\author[label1]{Zhaojian Li}
\author[label1]{Kaixiang Zhang}
\author[label1]{Dong Chen}
\author[label1]{Kyle Lammers}
\author[label2]{Renfu Lu}

\address{Zhaojian Li (lizhaoj1@msu.edu) is the corresponding author}
\address[label1]{Department of Mechanical Engineering, Michigan State University, East Lansing, MI 48824, USA}
\address[label2]{Department of Agriculture (USDA) Agricultural
Research Service (ARS), East Lansing, MI 48824, USA}

\begin{abstract}

Automated apple harvesting has attracted significant research interest in recent years due to its potential to revolutionize the apple industry, addressing the issues of shortage and high costs in labor.  One key technology to fully enable efficient automated harvesting is accurate and robust apple detection, which is challenging due to complex orchard environments that involve varying lighting conditions and foliage/branch occlusions. Furthermore, clustered apples are common in the orchard, which brings  additional challenges as the clustered apples may be identified as one apple. This will  cause issues in localization for subsequent robotic operations. In this paper, we present the development of a novel deep learning-based apple detection framework,   Occluder-Occludee Relational Network (O2RNet), for robust detection of apples in such clustered environments. This network exploits the occuluder-occludee relationship modeling head by introducing a feature expansion structure to enable the combination of layered traditional detectors to split clustered apples and foliage occlusions. More specifically, we collect a comprehensive apple orchard image dataset under different lighting conditions (overcast, front lighting, and back lighting) with frequent apple occlusions.  We then develop a novel occlusion-aware network for apple detection, in which a feature expansion structure is incorporated into the convolutional neural networks to extract additional features generated by the original network for occluded apples. Comprehensive evaluations are performed, which show that the developed O2RNet outperforms state-of-the-art models with a higher accuracy of 94\% and a higher F1-score of 0.88 on apple detection.

\end{abstract}

\begin{keyword}
computer vision, apple detection, fruit harvesting, occlusion-aware detection, transfer learning
\end{keyword}

\end{frontmatter}


\section{Introduction}
\label{sec:intro}
Driven by rising costs and growing shortages in harvesting labor, robotic apple harvesting has gained increased research attention over the past decade. In the U.S. alone, fruit harvesting requires more than \textit{10 million worker hours} annually, attributing to approximately $15\%$ of the total apple production cost \citep{Apple_stat}. Mechanization and automation promise next-gen harvesting systems with low operating cost and high efficiency, as well as the ability to assess individual fruit for quality and maturity at the point of harvest \citep{li2016design}.

As such, several research groups have been developing robotic harvesting systems \citep{kang2019fruit, wan2020faster, qingchun2012study, de2011design,ZHANG_ROBOT1}. Despite progresses, several important challenges in developing a fully functional robotic harvesting system remain, and no commercially-viable systems are yet available in the market. One key challenge that is pointed out by the existing works is efficient and robust fruit detection in the presence of varying light conditions and fruit/foliage occlusions. Indeed, the perception system provides the robot system with information on target fruits, which are first and foremost for subsequent planning and control tasks. 
In addition,  fruit perception techniques have also been used in other applications of interest, including yield estimation and  crop health status monitoring~\citep{patel2011fruit}. 
Perception in unstructured orchard environments, however, is a daunting task as a result of variations in illumination and appearance, noisy backgrounds, and clustered environments with occlusions \citep{chu2021deep}. The goal of this paper is thus to present a novel deep learning-based detection algorithm to convergently address the aforementioned challenges. We show that the developed algorithm is able to achieve state-of-the-art performance. Before describing the technical details, we review relevant backgrounds and state-of-the-art approaches to put our algorithm in better context. 


\subsection{Image Sensing Techniques}
Vision-based perception schemes can be classified into four categories based on the sensor used: monocular camera scheme, binocular stereovision scheme, laser active visual scheme, and thermal imaging scheme, which cover both two-dimension imaging schemes and three-dimension imaging schemes \citep{zhao2016review}. Specifically, the monocular scheme uses a single camera to acquire image data, and it is widely used in  fruit harvesting due to its low cost and rich information provided by the RGB images. For instance, \cite{tian2019apple} developed an improved YOLOv3 \citep{redmon2018yolov3} model based on a single camera to detect apples with an accuracy of 85.0\%. In \cite{kang2020fast}, the authors proposed a new LedNet model for apple detection that achieves an accuracy of 85.3\%. The main disadvantage of the monocular scheme is that the color images are sensitive to fluctuating illumination. 

Different from the monocular camera schemes, the binocular stereovision schemes exploit two cameras separated in a certain distance/angle to obtain two image data on the same scene. The point cloud of fruit can then be constructed through triangulation on extracted features \citep{sun2011fruits}. For instance, \cite{si2015location} used a stereo camera to detect and localize mature apples in tree canopies, and achieved an accuracy of 89.5\%. In \cite{xiang2014recognition}, the authors developed a clustered tomato detection method based on a stereo camera, and the recognition accuracy was 87.9\%. Although the stereovision scheme tends to render better results, it suffers from high complexity, long computation time, and uncertainties in  stereo matching \citep{hannan2004current}. 

On the other hand, the laser active visual schemes obtain three-dimensional features using laser scans, where laser beam reflections are exploited to generate a 3D point cloud based on the time-of-flight principle. The 3D point cloud can then be used to reconstruct the scene.
For example, \citep{tanigaki2008cherry} utilized infrared laser scanning devices to recognize cherry on the tree. \citep{zhang2015computer} acquired a total of 200 images for independent `Fuji' apples and developed an apple recognition method using the near-infrared linear-array structured
light for 3D reconstruction. \citep{tsoulias2020apple} proposed a point cloud based apple detection method using a LiDAR laser scanner and reached a 88.2\% overall accuracy on the defoliated tree dataset \citep{tsoulias2020apple}. Note the defoliated scene is significantly less challenging than the real orchard conditions during the harvest season. Furthermore, the laser point cloud is generally sparse and it is challenging to be used in real-world orchards with dense backgrounds. The high cost and complexity also limit its practical application in agricultural applications. 

Finally, the thermal imaging schemes make use of the distinct thermal characteristics of fruit and leaves (e.g., the different temperature distributions) to obtain the visualization of infrared radiation \citep{lu2014detecting}. In \cite{bulanon2008study},  citruses are successfully segmented using a thermal infrared camera according to the largest temperature difference in both day and night conditions. An enhanced approach for fruit detection \citep{bulanon2009image} was developed using the combination of the thermal image and the color image. The results showed a promising performance under weak lighting environments. However, in the thermal imaging scheme, the accuracy of recognition is largely affected by the shadow of the tree canopy \citep{stajnko2004estimation}. 

Considering the cost, performance, and real-time constraints, our work focuses on the monocular camera scheme, the state-of-art of which will be discussed next. 

\subsection{Recognition Approaches}
Image-based fruit recognition approaches can be classiﬁed into \textit{feature analysis} approaches and \textit{deep learning-based} approaches, depending on how features are obtained. In \textit{feature analysis} approaches, hand-crafted features are first extracted based on the fruit characteristics, and classification approaches are then developed to recognize fruit.  \cite{slaughter1987color, sites1988computer} developed thresholding methods to classify fruit from other background objects using smoothing filters that remove irrelevant noises. The large segmented regions are then recognized as fruits. This method is capable of segmenting fruit regions in simple backgrounds but it is susceptible to varying lighting conditions and complex canopies. \cite{whittaker1987fruit,benady1992locating} proposed a circular Hough Transform approach to obtain binary edge images and then used a voting matrix to identify fruits. This approach is sensitive to complex structured environments and it generally fails in a dense scene. In \cite{qiu1992maturity,cardenas1991machine,levi1988image,zhao2005tree},  they combined the shape and texture of the fruit to obtain a richer set of feature representations. Then, extracted features between fruit and leaves are compared and contrasted to identify the fruits. However, this method is also sensitive to lighting conditions and occlusions. 

On the other hand, deep learning-based approaches have found great successes in object detection and semantic image segmentation \cite{sa2016deepfruits, bargoti2017image}. They can learn feature representations automatically without the need of manual feature engineering. Compared to conventional methods,  Convolutional Neural Networks (CNNs) have been showing great advantages in the field of object detection in recent years. The CNN makes it possible to recognize fruits in complex situations due to its deep extraction of high-dimensional features of objects.  R-CNN and its variants Fast R-CNN and Faster R-CNN \citep{girshick2014rich, girshick2015fast, ren2015faster} have enjoyed particular successes. Their key idea is to first obtain regions of interest and then perform classification in the region. The Region proposal network (RPN) is employed to reduce high computational costs so that the model can simultaneously predict and classify object boundaries at each location. The parameters of the two networks are shared, which results in much faster inference and are thus optimized for real-time purposes.
Faster Region-Based CNN, proposed by \cite{sa2016deepfruits}, employed transfer learning using ImageNet, and used both early fusion and late fusion to integrate RGB and NIR (near infrared) inputs. 
Modified Inception-ResNet (MI-ResNet) \citep{rahnemoonfar2017deep} used deep simulated learning for yield estimation. The model was developed to address challenges including the varying degree of fruit sizes and overlap, natural lighting,  and foliage occlusions. The overhead for object detection and localization is optimized by utilizing synthetic data for training, and reaching the accuracy of 91\% on their  fruit dataset. 
You Only Look Once (YOLOv3) \citep{redmon2018yolov3}, a representative of the one-stage object detector, detects the fruit on the entire image and classifies fruit variety into uncertainty retail conditions without the help of RPN. Specifically, YOLOv3 uses logistic regression to predict an objectless score for each bounding box. Due to the simple optimization pipeline, YOLOv3 enjoys much faster inference than the aforementioned region-based methods.
EfficientDet \citep{tan2020efficientdet}, an augmented variant of YOLOv3, exploits a pyramid network to enable the detection of scaling targets. 

However, the aforementioned Deep CNN approaches do not address the challenge of fruit/foliage occlusions in real-world orchards. Towards that end, Compositional Convolutional Neural Network (CompositionalNet) \citep{kortylewski2020compositional} was proposed to detect partially occluded objects. The framework exploits a differentiable fully compositional model that uses occluder kernels to localize occluders (the occluding objects). Bilayer Convolutional Network (BCNet) \citep{ke2021deep}, another model to address the occlusion challenge, applies two Graph Convolutional Network (GCN) layers to separately infer the occluding objects (occluder) and partially occluded instance (occludee). By sharing parameters between  the top and bottom GCN layers, BCNet decouples the occluder and occludee on the input image. Superior performance was reported on occluded scenarios. 

\subsection{Our Contributions}
In this paper, we develop a novel Occluder-Occludee Relational Network (O2RNet) to enhance apple detection in the presence of occlusions in clustered apples that are frequently present in real-world orchards. Specifically, we employ ResNet \citep{he2016deep} and RPN \citep{ren2015faster} to extract features of targets and utilize occluder-occludee layers to split candidates into occluder and occludee. Compared to other occlusion models, we only use bounding boxes as labels instead of pixel-level masks that contain more texture and shape information. In addition, we present a new apple dataset\footnote{The database is open-sourced at \url{https://github.com/pengyuchu/MSUAppleDatasetv2.git}.} collected in two Michigan apple orchards in multiple harvesting seasons. We evaluate the performance against state-of-the-art object detection models and demonstrate superior performances.  The contributions of this paper are highlighted as follows:

\begin{enumerate}
\item The presentation of a comprehensive apple dataset consisting of $900$ images with different lighting conditions and occlusion levels collected in multiple orchards across multiple harvesting seasons.
\item The development of Occluder-Occludee Relational Network (O2RNet), a novel occlusion-aware network for enhanced apple detection in the presence of occlusion due to apple clusters. 
\item A comprehensive evaluation and benchmark of $12$ state-of-the-art deep learning-based models for apple detection where we show that the developed O2RNet outperforms state-of-the-art algorithms.
\end{enumerate}


\section{Materials and Methods}
\label{sec:datapre}
\subsection{Data Collection and Processing}
\label{sec:datapre}
\begin{figure}[!h]
\centerline{\includegraphics[width=0.95\columnwidth]{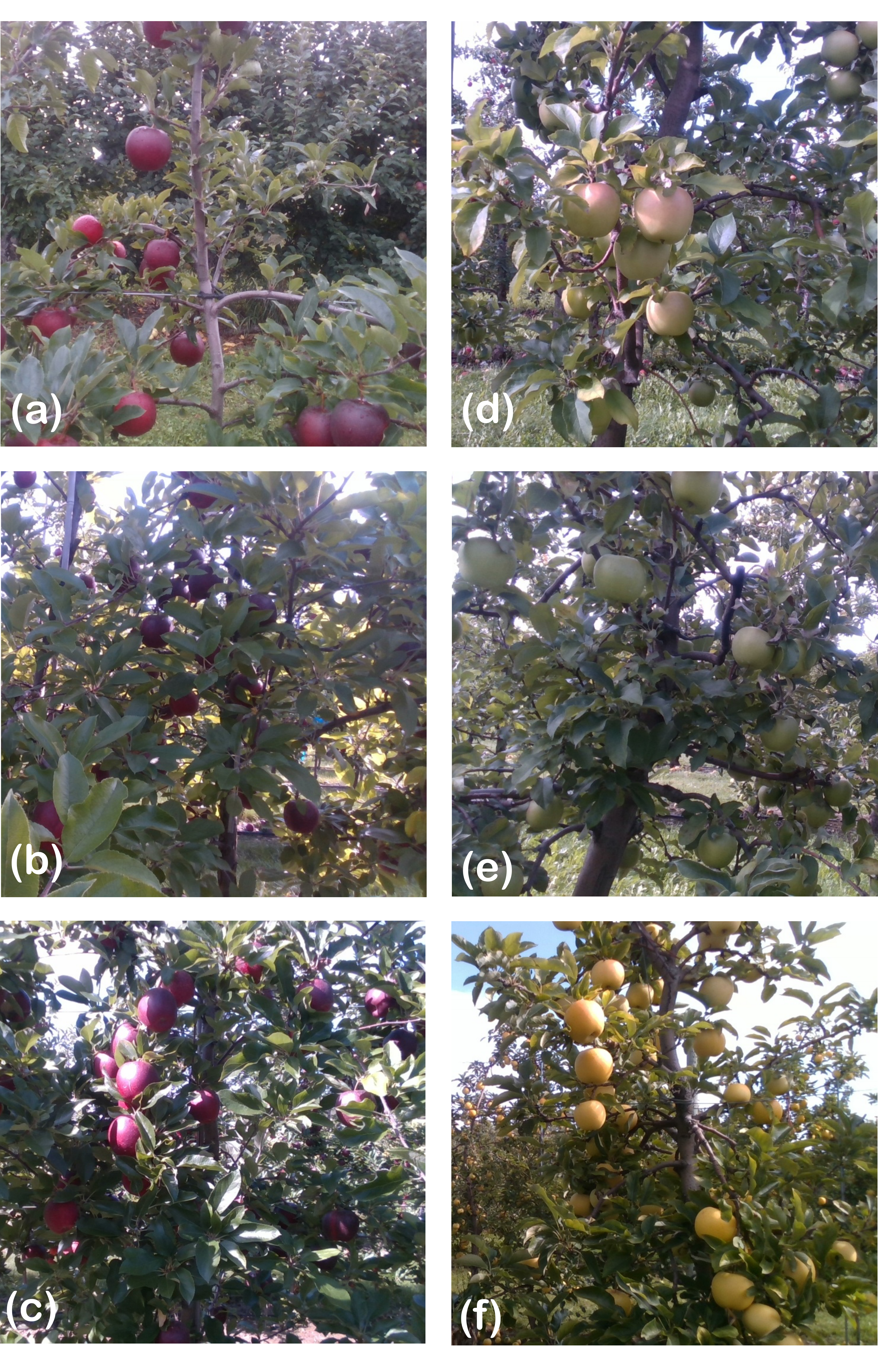}}
\caption{Six sample images from the collected dataset: (a)-(c) apples on older trees under overcast, back-lighting, and direct lighting conditions, respectively; and (d)-(e) apples on younger trees under overcast, back-lighting, and direct lighting conditions, respectively.}
\label{fig:apple}
\end{figure}

In this study, apple images were taken in two  orchards: the commercial orchard in Sparta, Michigan, USA during the 2019 harvest season and the experimental orchard of Michigan State University in East Lansing, Michigan, USA during the 2021 harvest season. The apples are mainly `Gala' that are generally red over a green/yellow background (see Fig.~\ref{fig:apple}). An RGB camera with a resolution of $1280 \times 720$ was used to take images of apples at a distance of $1-2$ meters from the tree trunks, which is the typical range of harvesting robots \citep{de2011design,ZHANG_ROBOT1,Zhang_ROBOT2}. The images were collected across multiple days to cover both cloudy and sunny weather conditions. In a single day, the data were also collected at different times of the day, including 9am, noon, and 3pm, to cover different lighting angles: front-lighting, back-lighting, side-lighting, and scattered lighting. Furthermore, we also captured clustered apples with different occlusion levels including both foliage and branches occlusion. When capturing images, the camera was placed parallel to the ground and directly facing the trees to mimic the harvesting scenario. Compared to our previous work \citep{chu2021deep}, an additional set of $200$ images were added to extend our dataset to a total of $900$ images where a few sample images are shown in Fig.~\ref{fig:apple}. 

We then processed the acquired raw orchard images into formats that can be used to train and evaluate deep networks. Specifically, apples in the images were annotated by rectangles using VGG Image Annotator \citep{dutta2019vgg}, and the annotations were then compiled into the human-readable format. Compared to polygon and mask annotations, rectangular annotation used here accelerates data preparation, particularly in dense images like our dataset. The annotated dataset was then split into training, validation, and test subsets with the apple quantities of $7522$, $3001$, and $3995$ respectively. The processed image database is open-sourced and can be accessed at \url{https://github.com/pengyuchu/MSUAppleDatasetv2.git}.

\subsection{Transfer Learning}
\label{subsec:tl}
We employ transfer learning to enable faster training and improved performance.  Transfer learning is a popular scheme that starts the model development with a pre-trained model on a large-scale dataset and then fine-tunes the model on a customized dataset from the specific domain of interest \citep{zhuang2020comprehensive}. For apple detection in this study, we used ImageNet \citep{deng2009imagenet} to pre-train each model and only replaced the last fully-connected layers in each model. Since there are objects of apple and alike in ImageNet, the pre-trained models converge faster in our customized apple dataset compared to randomized initial parameters.

\subsection{Performance Metrics}
For model development and evaluation, conventionally the apple dataset is randomly partitioned into training, validation, and test sets for model training and evaluation, respectively. To quantitatively evaluate the detection performance, we use performance metrics including precision, recall, and F1-score for algorithm evaluation. All detection outcomes are divided into four types: true positive ($TP$), false positive ($FP$), true negative ($TN$), and false negative ($FN$), based on the relation between the true class and predicted class. The precision ($P$) and recall ($R$) are defined as follows: 

\begin{equation}
    P = \frac{TP}{TP+FP}, \quad
    R = \frac{TP}{TP+FN}.
\end{equation}

The F1-score is then subsequently defined as:

\begin{equation}
    F1 = \frac{2\cdot P\cdot R}{P+R}.
\end{equation}

To better evaluate the precision between the prediction and the ground truth, we also employ Microsoft Common Objects in Context (COCO) dataset \citep{lin2014microsoft} evaluation metrics. Specifically, after the calculation of precision and recall, we  calculate the average precision ($AP$) and average recall ($AR$) based on different Intersection over Union (IoU) between the prediction and the ground truth. For example, $AP_{IoU=.50}$ or $AP_{50}$ denotes that AP is averaged over $IoU = 0.50$ values, which belongs to PASCAL VOC metric \citep{rezatofighi2019generalized}. We also use $AP_{IoU=.75}$ or $AP_{75}$, which is a stricter metric for model evaluations. In our study, we use a spectrum of $10$ IoU thresholds ranging $0.50:0.05:0.95$ to average over multiple IoUs to  obtain a comprehensive set of results.

\subsection{Data Augmentation}
Data augmentation is a method that can be adopted to increase data diversity for achieving robust training and enhanced performance of computer vision models. For example,  transformations and rotations are frequently employed to increase the number of images from a single source. It has been shown to be a powerful tool in agriculture applications \citep{wu2020dcgan, su2021data, divyanth2022image} as it generates additional data from existing orchard data. This is especially useful for applications with a limited dataset by detecting anomalies in images with different transformations and making it possible to generate new training examples without actually acquiring  new data.

Specifically, in the considered application of apple detection in orchards, the collected dataset can only cover a limited set of scenarios. Therefore, we applied several data augmentation techniques \citep{chlap2021review} on the collected and processed data to enhance the data diversity for improving the inference performance of our models. Specifically, besides geometric transformations including scaling, translating, rotating, reflecting, and shearing, we also applied color space augmentations such as modifying the brightness and contrast to fit different intensities. In addition,  we injected Gaussian noises on the collected images by randomly modifying the pixel intensities based on a Gaussian distribution. Furthermore, we applied Mixup by randomly selecting two images from the dataset and blending the intensities of the corresponding voxels of the two images \citep{lu2022generative}. Filtering is another augmentation approach we applied where we modify the intensities of each pixel using convolution \citep{shorten2019survey}. Specifically, we exploited sharpening \citep{shorten2019survey} to detect and intensify the edges of objects found within the image. We applied these additional augmentation techniques on our dataset and the benefits of data augmentation will be demonstrated in the experiment section.

\section{Methodology}
\label{sec:methodology}
In this section, we ﬁrst present the key challenges of object detection in clustered environments and an overview of the general object detection framework. Based on those, we describe the proposed Occluder-Occludee Relational Network (O2RNet) with explicit occluder-occludee relation modeling. Finally, we specify the objective functions for the entire network optimization, followed by details on the training and inference processes.

\subsection{Challenge and Main Idea}
\begin{figure}[H]
\center
\includegraphics[width=0.98\columnwidth]{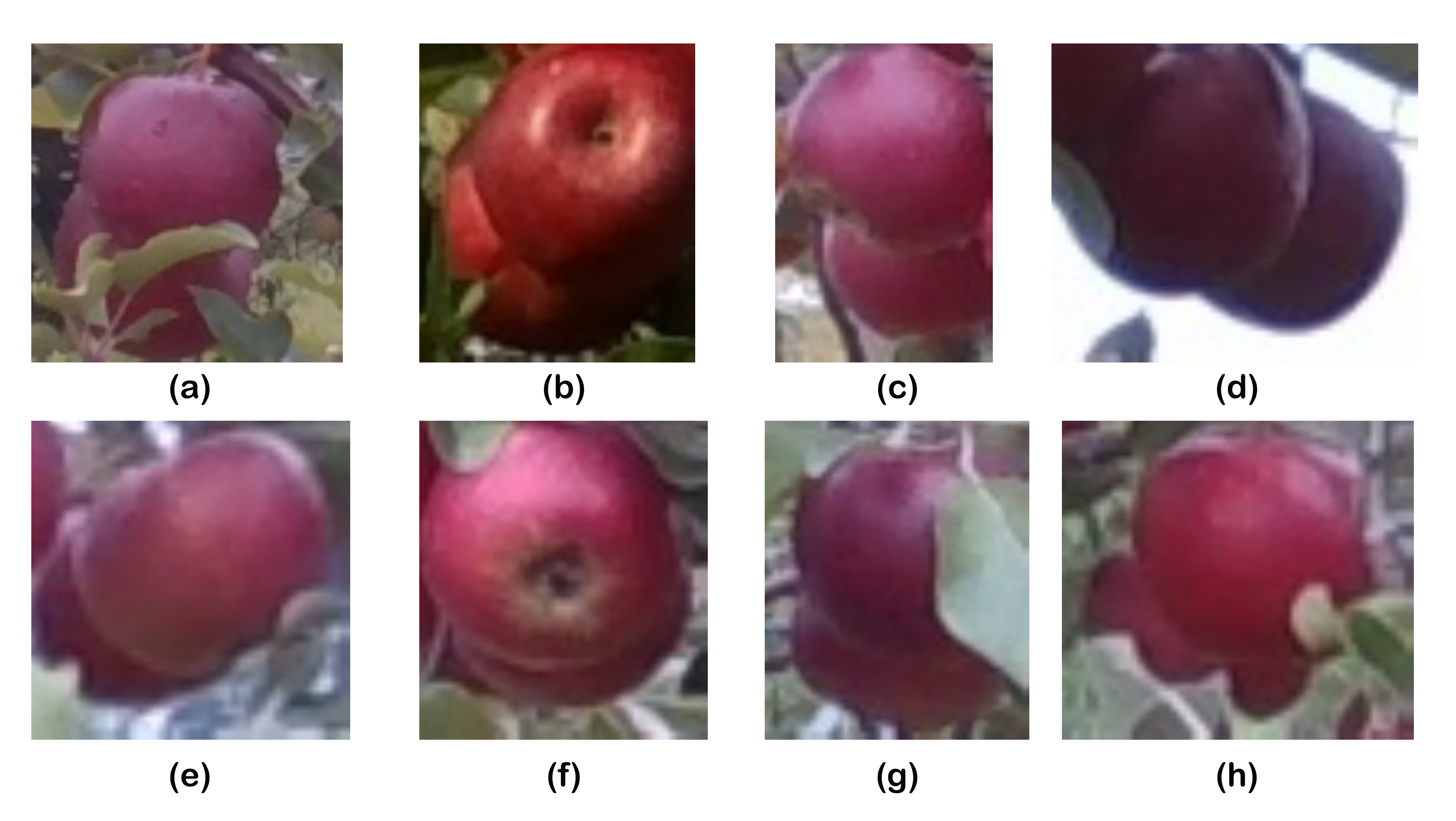}
\caption{Eight sample images from the collected dataset show cascaded apples in different occlusion levels: (a)-(d) apples are in the normal occlusion and can be identified in most models; (e)-(h) apples are highly cascaded and usually detected as one apple.}
\label{fig:occlu_case}
\end{figure}
\begin{figure}[H]
\centerline{\includegraphics[width=0.85\columnwidth]{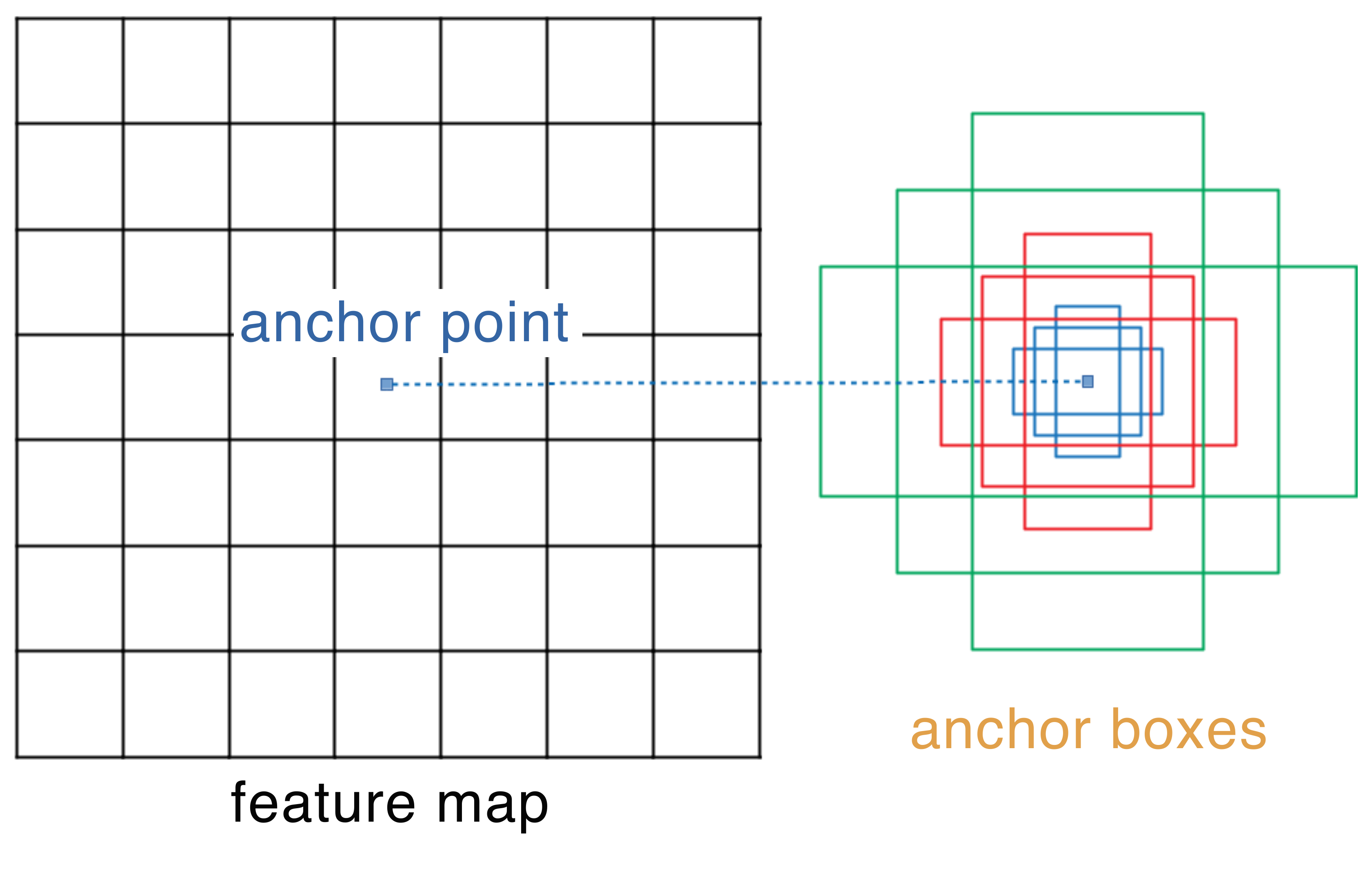}}
\caption{Illustration of how RPN works: The RPN selects anchor points on the feature map and generates anchor boxes for each anchor point. The anchor boxes are generated based on two parameters — scales and aspect ratios.} 
\label{fig:rpn_results}
\end{figure}

For images with heavy occlusions, multiple overlapping objects captured in the same bounding box can result in confusing object outlines from both front objects and occlusion boundaries. In apple orchards, the apple clusters are very common (see Fig.~\ref{fig:occlu_case} for a few examples). However,  the prediction head design of Faster R-CNN directly regresses the occludee with a fully convolutional network, which neglects both the occluding instances and the overlapping relations between objects. With this limitation, Faster R-CNNs will inevitably omit some occludes due to Non-maximum Suppression (NMS). On the other hand,  with a properly tuned threshold, the RPN can propose many candidates after feeding the target features from CNN (see  Fig.~\ref{fig:rpn_results}), but the NMS will suppress the nearby bounding boxes and neglect occludees. Motivated by this observation, the proposed O2RNet aims at extending the existing two-stage object detection methods by adding an occlusion perception branch parallel to the original object prediction pipeline. By explicitly modeling the relationship between occluder and occludee, the interactions between objects within the Region of Interest (RoI) region can be well incorporated during the bounding box regression stage.

\subsection{O2RNet Workflow}
\begin{figure*}[t]
\centerline{\includegraphics[width=0.9\textwidth,height=6.5cm]{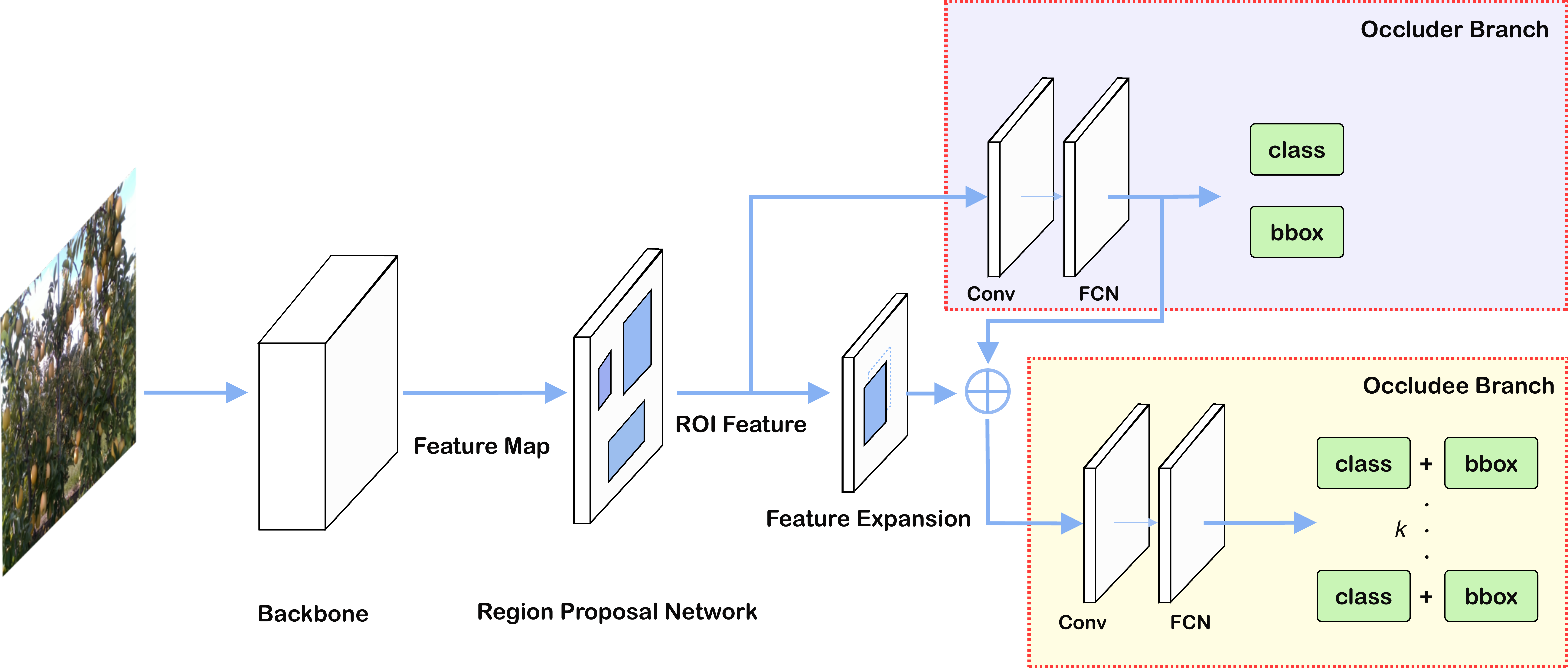}}
\caption{Network structure of the proposed Occluder-Occludee Relational Network (O2RNet). It consists of a feature learning backbone, RoI feature extraction, and object detection heads with  occluder and occludee branches. The Feature Expansion Structure (FES) provides expanded RoI features along with features from the occluder branch to facilitate the detection of occludee.}
\label{fig:cnn}
\end{figure*}

In this subsection, we describe our proposed O2RNet. As illustrated in Fig.~\ref{fig:cnn}, the O2RNet follows the two-stage architecture used in Faster R-CNN \citep{ren2015faster} and  consists of three main parts. First, we use a Residual Network (ResNet) \citep{he2016deep} as the backbone for feature learning/extraction over the entire image. Specifically, we instantiate ResNet-101-FPN \citep{he2017mask} as its backbone for feature extraction, as it outperforms other single ConvNets mainly due to its capability of maintaining strong semantic features at various resolution scales. Even though ResNet101 is a deep network, the residual blocks and dropouts function help it avoid gradient vanishing and exploding problems. Second, we employ an RPN \citep{ren2015faster} to generate object regions, which is a small convolutional network to convert feature maps into scored region proposals around where the object lies. The generated proposals with a certain height and width are called anchors, which are a set of predefined bounding boxes. The anchors are designed to capture the scale and aspect ratio of specific object classes and are typically chosen to be consistent with object sizes in the dataset.  RPN is mainly used for predicting bounding boxes in Faster R-CNN but it can also provide enough anchors with different scales that will be exploited in our network as explained in the sequel. Third, we build an occlusion-aware modeling head with a structure of two classification and regression branches for occluder and occludee for decoupling overlapping relations and segments the instance proposals obtained from the RPN. Compared to the traditional class-agnostic classification, we divide this task into two complementary tasks: occluder prediction using the original classification head and occludee modeling with an additional Feature Expansion Structure (FES), where the occluder predictions provide rich foreground cues like textures and the FES predicts the positions of occluding regions to guide occludee object regression.

More specifically, an input image is first processed by the ResNet backbone to extract intermediate convolutional features for downstream processing. The object detection head (i.e., RPN) then predicts bounding box proposals, which are then consumed by the occlusion perception branches into the occluder branch and the occluee branch. For the occluder branch, we adopt the object detection head in Faster R-CNN \citep{ren2015faster} to output positions as well as categories for instance candidates and prepare the cropped RoI features for the occludee branch. In the occludee branch, the input consists of both cropped RoI features from the occluder branch and expanded features from FES, which is targeted for modeling occluded regions by jointly detecting boundaries. Essentially, the distilled occlusion features are added to the original input RoI features and passed to the next module. Finally, the occludee branch, which has a similar structure to the occluder branch, predicts the occludee guided by these expanded features and outputs classes and bounding boxes for the partially occluded instances. We next describe the occluder-occludee relational modeling in more details.

\subsection{Occluder-Occludee Relationship Modeling}

For highly-overlapped apples, in typical Faster-RCNN-based models, the generated region proposals corresponding to the partially occluded ones may be separated into disjoint subregions by the occluder. As such, we employ the FES to obtain boundary features from the occludee, where expansion in each direction extends the potential proposals for the occludee. In our implementation, we expand $t$ steps in $k$ ($k=8$ in this study) directions from the original RoI proposals, and the expanded RoI proposals will contain additional boundary features. The rationale is that irregular occlusion boundaries unrelated to the occludee can cause confusion to the network, which in turn provides essential cues for decoupling occludees from occluders. Therefore, we explicitly model occlusion patterns by detecting bounding boxes of the occluders using the occluder detection branch, and since the occludee detection branch jointly predicts bounding boxes for the occludee, the overlap between the two layers can be directly identiﬁed as occlusion boundary that can thus be distinguished from the real object bounding boxes. In order to reach this goal, the occluder modeling module is designed as a simple $3 \times 3$ convolutional layer followed by one FCN layer, the output of which is fed to the up-sampling layer and one $1 \times 1$ convolutional layer to obtain one channel feature map for occludee branch. 


\subsection{End-to-end Learning}
As we have two separate detection heads in the occluder and the occludee branches, we define two loss functions in the following way. For the occluder branch, we adopt the loss function used in Faster R-CNN \citep{ren2015faster}, which defines a multi-task loss on each sampled region of interest as
\begin{equation}
L_{Occluder} =  L_{cls} + L_{bbox},\\
\end{equation}
where $L_{cls}$ and $L_{bbox}$ are, respectively, classification loss and bounding box loss defined in Faster R-CNN \cite{ren2015faster}.

The final loss $L$ is a weighted sum of the loss from occluder branch and the loss from occludee branch defined as:
\begin{equation}
L = \lambda_1 L_{Occluder} + \lambda_2 L_{Occludee}.\\
\end{equation}
Here $L_{Occludee}$ is the occludee branch loss that is the sum of the $k$ expanded proposal losses, i.e.,
\begin{equation}
L_{Occludee} =  \sum_{i=0}^k (L^i_{cls} + L^i_{bbox}).\\
\end{equation}
Here $\lambda_1$ and $\lambda_2$ are two positive linear weights and $\lambda_1+\lambda_2=1$, which are tuned to balance the two loss functions. In our study, $\lambda_1$ was tuned to be $\{1.0, 0.75, 0.5, 0.25, 0 \}$ on various trials for cross-validation.

\subsection{Training and Inference}

During the training process, we ﬁlter out parts of the non-occluded RoI proposals to keep occlusion cases taking up 50\% for balanced sampling. SGD with momentum is employed to train the model with $60K$ iterations where it starts with $1K$ constant warm-up iterations. The batch size is set to 2 and the initial learning rate is $0.01$ with a weights decay of $0.95$. In our study, ResNet-101-FPN is used as the backbone and the input images are resized without changing the aspect ratio, i.e., by keeping the shorter side and longer side of no more than $1200$ pixels.
For inference, the occludee branch predicts bounding boxes for the occluded target object in the high-score box proposals generated by the RPN, while the occluder branch produces occlusion-aware features as input for the occludee branch. The one with the highest score is then chosen as the output.

\begin{table*}[!ht]
\renewcommand{\arraystretch}{1.2}
\centering
\fontsize{6}{7}
\selectfont
\caption{Performance of O2RNet on the customized apple dataset. The step is from FES, which represents how much features expanded. The evaluation uses AP, AR, and F1-score at the different IoUs.}
\label{tab:oornet_eval}
\resizebox{0.8\textwidth}{!}{
\begin{tabular}{lcccccccc}
\hline
Model                   & Step & $AP$    & $AP_{50}$ & $AP_{75}$ & $AR$    & $AR_{50}$ & $AR_{75}$ & F1-Score \\ \hline \hline
\multirow{3}{*}{O2RNet} & \textbf{t=1}  & \textbf{0.511} & \textbf{0.945}  & \textbf{0.935}  & \textbf{0.351} & \textbf{0.938}  & \textbf{0.803}  & \textbf{0.864}    \\  
                        & t=2  & 0.490 & 0.920  & 0.900  & 0.330 & 0.900  & 0.770  & 0.820    \\  
                        & t=3  & 0.490 & 0.920  & 0.904  & 0.328 & 0.900  & 0.770  & 0.820    \\ \hline
\end{tabular}}
\end{table*}

\begin{table}[!h]
    \renewcommand{\arraystretch}{1.2}
    \centering
    \fontsize{6}{7}
    \selectfont
    \caption{Model parameters numbers between the state-of-the-art networks and our proposed Occluder-occludee Relational Network (O2RNet). ``M'' stands for a million.}
    \resizebox{0.38\textwidth}{!}{
    \begin{tabular}{l c}\hline
        Models & Parameters  \\
        \hline \hline
        FCOS & 2.0M \\
        YOLOv4 & 0.6M \\
        Faster R-CNN (ResNet50) & 2.0M\\
        Faster R-CNN (ResNet101) & 3.6M \\
        EfficientDet-b0 & 0.1M  \\
        EfficientDet-b1 & 0.3M  \\
        EfficientDet-b2 & 1.2M  \\
        EfficientDet-b3 & 1.6M  \\
        EfficientDet-b4 & 2.4M  \\
        EfficientDet-b5 & 3.6M  \\
        CompNet via BBV & 0.8M \\
        CompNet via RPN & 1.4M \\
        O2RNet (ResNet50) & 2.0M \\
        O2RNet (ResNet101) & 3.6M \\
        \hline
    \end{tabular}}
    \label{tab:models}
\end{table}

\section{Experiment and Discussions}

\subsection{Experimental Setup}

In this section, we evaluate the efficacy of the proposed O2RNet on the processed data as discussed in Section~\ref{sec:datapre}. The network hyper-parameters, including the momentum, learning rate, decay factor, training steps, and batch size, are set as $0.9$, $0.001$, $0.0005$, $934$, and $1$, respectively, through cross-validation. The input image size is $1280 \times 720$, which is aligned with the resolution of the camera used in our data collection. To better analyze the training process, we set up $80$ epochs for training. We exploit a pre-trained model on the COCO dataset \citep{lin2014microsoft}, where we train on 2017train ($115k$ images) and evaluate results on both 2017val and 2017test-dev to pre-train model parameters. This pre-trained model generally only takes $50$ epochs to converge. By tuning the steps $t$ in FES, different results are obtained and listed in Table~\ref{tab:oornet_eval}, which shows that O2RNet with $t=1$ leads to the best performance.


\begin{table*}[!ht]
\renewcommand{\arraystretch}{1.2}
\centering
\caption{Performance of O2RNet on the augmented dataset. The geometric transformations consist of rotation, flipping and scaling. The color space transformations consist of brightness and contrast shifting. Finally, all of the augmentation methods are integrated to evaluate the O2RNet.}
\label{tab:aug_results}
\resizebox{0.8\textwidth}{!}{
\begin{tabular}{lcccccccc}
\hline
Augmentation                    & $AP$    & $AP_{50}$ & $AP_{75}$ & $AR$    & $AR_{50}$ & $AR_{75}$ & F1-Score \\ \hline \hline
Base  & 0.51 & 0.92  & 0.90  & 0.35 & 0.91  & 0.80  & 0.84   \\ 
Geometric transformations (GTs)  & 0.52 & 0.93  & 0.91  & 0.35 & 0.91  & 0.80  & 0.85    \\  
Color space transformations (CSTs)  & 0.52 & 0.93  & 0.91  & 0.35 & 0.91  & 0.81  & 0.85    \\  
Gausian noise & 0.48 & 0.91  & 0.90  & 0.34 & 0.91  & 0.80  & 0.83    \\ 
Mixup  & 0.52 & 0.93  & 0.92  & 0.35 & 0.92  & 0.81  & 0.85    \\ 
Sharpening  & 0.52 & 0.92  & 0.90  & 0.35 & 0.91  & 0.80  & 0.84    \\ 
GTs$+$CSTs$+$Mixup  & \textbf{0.52} & \textbf{0.96}  & \textbf{0.94}  & \textbf{0.36} & \textbf{0.94}  & \textbf{0.83}  & \textbf{0.88}    \\ 
All  & 0.52 & 0.94  & 0.92  & 0.36 & 0.92  & 0.83  & 0.86    \\ \hline
\end{tabular}}
\end{table*}

\begin{table*}[!ht]
\caption{Performance comparison of our own models and other $12$ state-of-the-art deep learning models on the customized apple dataset.}
\renewcommand{\arraystretch}{1.2}
\centering
\resizebox{0.8\textwidth}{!}{
\begin{tabular}{llccccccc}
\hline
\multicolumn{2}{l}{Models} & $AP$ & $AP_{50}$ & $AP_{75}$ & $AR$ & $AR_{50}$ & $AR_{75}$ & F1-score\\ \hline \hline
\multicolumn{2}{l}{FCOS \citep{ahmad2021performance}}                                    & 0.48  & 0.89 & 0.87 & 0.34 & 0.87 & 0.78 & 0.80  \\ \hline
\multicolumn{2}{l}{YOLOv4 \citep{pandey2021autonomy}}                                   & 0.45  & 0.87 & 0.84 & 0.29 & 0.84 & 0.73 & 0.76 \\ \hline
\multicolumn{1}{l}{\multirow{2}{*}{Faster R-CNN}}       & ResNet50 \citep{norsworthy2012reducing}       & 0.48  & 0.89 & 0.87 & 0.32 & 0.87 & 0.78 & 0.81  \\   
\multicolumn{1}{l}{}                              & ResNet101 \citep{norsworthy2012reducing}           & 0.49  & 0.94 & 0.93 & 0.31 & 0.84 & 0.75 & 0.82         \\ \hline
\multicolumn{1}{l}{\multirow{6}{*}{EfficientDet}} & EfficientDet-b0 \citep{oerke2006crop}  & 0.45  & 0.89 & 0.85 & 0.30 & 0.82 & 0.71 & 0.77         \\   
\multicolumn{1}{l}{}   & EfficientDet-b1 \citep{oerke2006crop}                    
& 0.45  & 0.89 & 0.86 & 0.30 & 0.82 & 0.72 & 0.77         \\ 
\multicolumn{1}{l}{}   & EfficientDet-b2 \citep{oerke2006crop}                       
& 0.46  & 0.89 & 0.87 & 0.30 & 0.82 & 0.73 & 0.78 \\  
\multicolumn{1}{l}{}   & EfficientDet-b3 \citep{oerke2006crop}  
& 0.49  & 0.93 & 0.91 & 0.32 & 0.84 & 0.75 & 0.81 \\  
\multicolumn{1}{l}{}   & EfficientDet-b4 \citep{oerke2006crop}                       
& 0.50  & 0.94 & 0.92 & 0.34 & 0.88 & 0.78 & 0.82 \\   
\multicolumn{1}{l}{}   & EfficientDet-b5 \citep{oerke2006crop}         
& 0.50  & 0.95 & 0.93 & 0.34 & 0.88 & 0.78 & 0.83 \\ \hline
\multicolumn{1}{l}{\multirow{2}{*}{CompNet}}    & CompNet via BBV \citep{young2014future}  
& 0.50  & 0.94 & 0.92 & 0.36 & 0.94 & 0.80 & 0.85 \\   
\multicolumn{1}{l}{}                              & CompNet via RPN \citep{fennimore2019robotic} 
& 0.51  & 0.95 & 0.94 & 0.35 & 0.94 & 0.80 & 0.86 \\ \hline

\multicolumn{1}{l}{\multirow{2}{*}{O2RNet}}    & O2RNet-ResNet50   
& 0.50  & 0.93 & 0.91 & 0.35 & 0.91 & 0.80 & 0.84 \\  
\multicolumn{1}{l}{}                              & \textbf{O2RNet-ResNet101}             
& \textbf{0.52}  & \textbf{0.96} & \textbf{0.94} & \textbf{0.36} & \textbf{0.94} & \textbf{0.83} & \textbf{0.88} \\ \hline
\end{tabular}}
\label{tab:evaluations}
\end{table*}

\subsection{Performance Comparison and Analysis}
\label{subsec:model_performance}

\begin{figure}[H]
\centerline{\includegraphics[width=1\columnwidth]{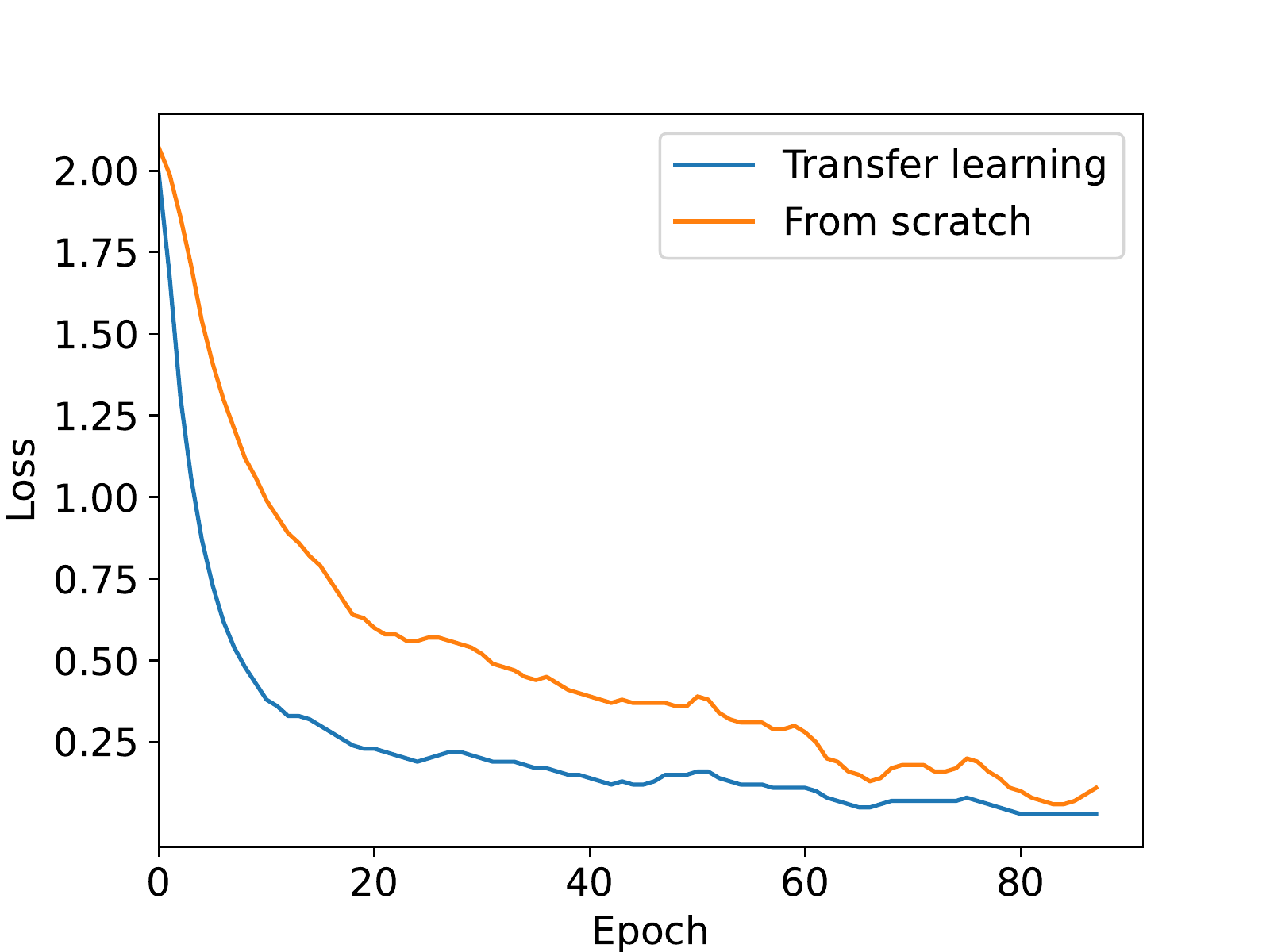}}
\caption{Training loss comparison between transfer learning and training from scratch on our model (O2RNet). The training loss with transfer learning from ImageNet apparently decreases and converges faster as compared with training from scratch.}
\label{fig:transfer_traincurve}
\end{figure}

To accelerate the model training on our customized dataset, we initialize parameters by transfer learning from ImageNet \citep{deng2009imagenet}. ImageNet provides large-scale images in different fields (including apples) and large-scale ground truth annotation. During the transfer learning process, our model learns specific characteristics with an effective transfer of features from ImageNet. Compared to randomized parameters, the results (see Fig.~\ref{fig:transfer_traincurve}) shows that our model converges faster as benefited from the pretraining on a large-scale database. 

Furthermore, data augmentation is another useful technique to optimize detection performance without increasing inference complexity. We applied five augmentation strategies, including geometric transformations (GTs), color space transformations (CSTs), Gaussian noise injection, mixup and sharpening data augmentation, to extend our dataset. The results are summarized in Table~\ref{tab:aug_results}. It shows that GTs such as rotation, flipping and scaling -- by changing the pixel position of the image and reordering apples in the image -- improve the accuracy performance by around $1\%$. Through changing color illumination and intensity of an image, CSTs also roughly increases the performance by $1\%$. Due to the sparsity of apples on some images, mixup helps enlarge apple density on the image and enhances the accuracy by $2\%$. It turns out that Gausian noise and sharpening do not help much, as they try to change textures and increase complexities on the dataset, which generate  confusing data and is not suitable for our model. 
Finally, the augmentation combination of GTs, CSTs and Mixup offers the best enhancement by increasing the accuracy of $4\%$ on our dataset.

To better evaluate the performance of our model, we compare our O2RNet with the-state-of-art object detection methods on our customized apple dataset (see Table~\ref{tab:models} for a list of benchmark models and their number of parameters). In particular, FCOS and YOLOv4 are representatives of one-stage detectors, achieving consistent improvement and demonstrating their effectiveness by outperforming the SSD method \citep{liu2016ssd} on several public datasets \citep{tian2019fcos, bochkovskiy2020yolov4}. We also evaluate Faster R-CNN and EfficientDet since they are state-of-the-art models with promising performance demonstrated in fruit harvesting-related works \citep{mekhalfi2021contrasting, yan2021real}. We also compare O2RNet with the state-of-the-art occlusion-aware network CompNet \citep{fennimore2019robotic}.

\begin{figure*}
\centerline{\includegraphics[width=0.98\textwidth]
{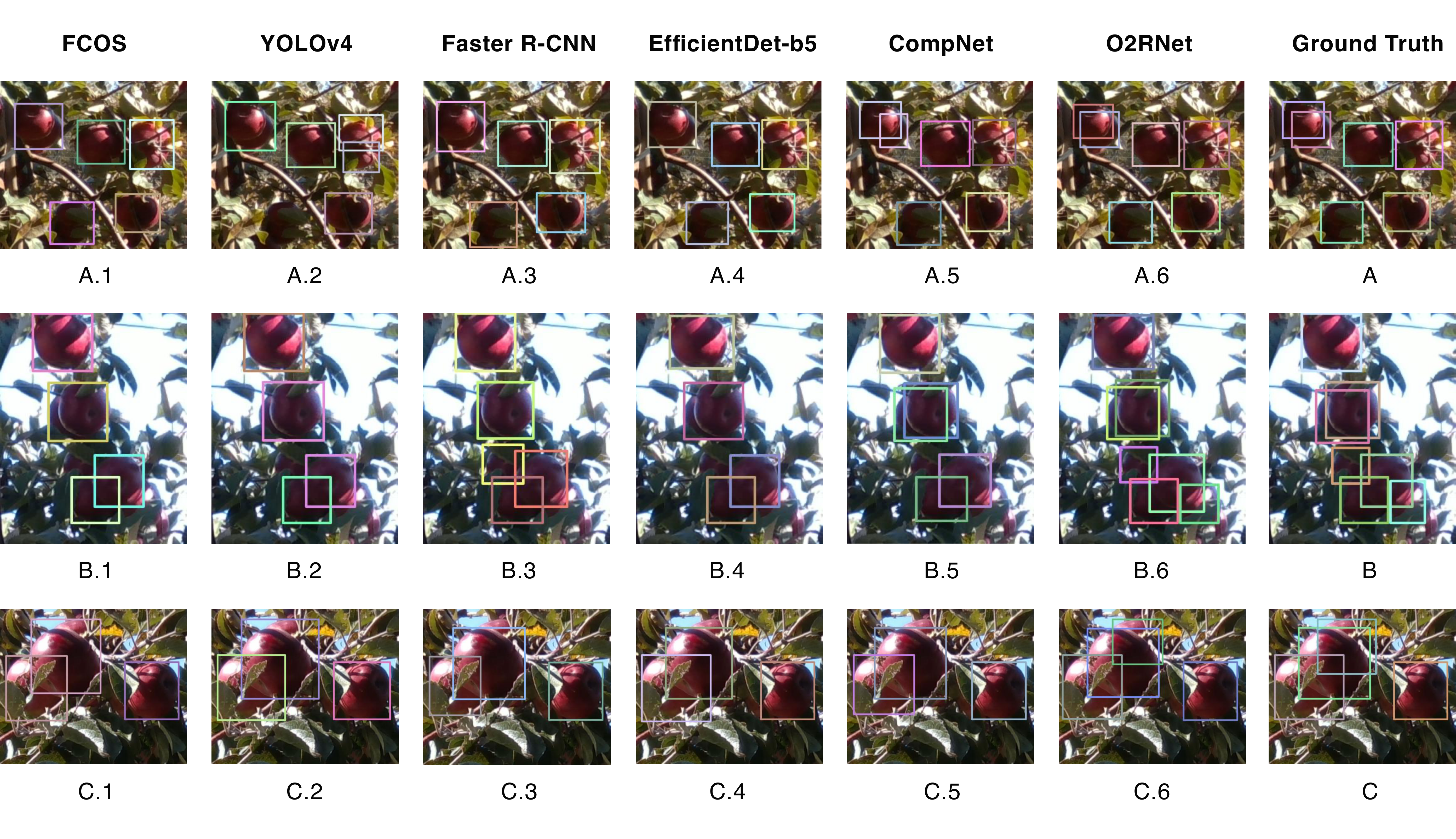}}
\caption{Results from six models on the various lighting conditions and occlusions}
\label{fig:results}
\end{figure*}

 We then use the same experimental setup to train each model and evaluate them on the same apple test dataset. The results are shown in Table~\ref{tab:evaluations}, which compares the detection precision and recall over different IoUs among the 14 selected models (including our O2RNet). Notably, in addition to FCOS, EfficientDet-b5 and Faster R-CNN achieved decent F1-scores of $0.83$ and $0.82$, respectively. Two occlusion-aware networks, CompNet and our O2RNet clearly outperform all traditional models with F1-scores of $0.86$ and $0.88$, respectively, and O2RNet clearly shows superior performance over CompNet. Some representative inference results are shown in Fig.~\ref{fig:results}. It can be seen that our O2RNet can effectively separate clustered apples and thereby improves the precision and recall and subsequently the F1-score.

\section{Conclusion}
\label{sec:conclu}

In this study, we collected a comprehensive apple dataset under different lighting conditions and at various occlusion levels from two real orchards. A novel Occluder-Occludee Relational Network (O2RNet) was developed to robustly detect clustered apples  from the dataset. Our developed O2RNet significantly reduced false detection and improved the detection rate by embedding relationships between the occluder and the occludee. State-of-art performance was demonstrated in  comprehensive experiments. We also found that transfer learning and data augmentation techniques were useful tools to enhance learning efficiency and model performance.

Our future work will include the incorporation of foliage information in the network design to further improve the detection performance since the current work only focuses on the clustered apples.  Furthermore, branch detection will be developed to provide necessary contextual information for the robot to maneuver, e.g., avoiding collisions with tree branches. Lastly, we will also investigate whether artificial lighting augmentation can enhance the detection performance.




\section*{Acknowledgement}
This research was funded by the USDA-ARS inhouse project 5050-43640-003-000D. The authors would also like to acknowledge Michigan State University’s Teaching and Research Center in Holt, Michigan and Schawllier’s Country Basket in Sparta, Michigan for their support for collecting the image data in the orchards. The authors thank Mr. Lingxuan Hao ans Wanqun Yang for helping label the orchard images.

\typeout{}
\bibliography{ref}
\end{document}


\title{Supplementary Materials: O2RNet: Occluder-Occludee Relational Network for Robust Apple Detection in Cluttered Environment} 
\author{Jack, Jack, Jack, Jack}
\date{\today} 
\maketitle 


\begin{table*}[!ht]

\renewcommand{\arraystretch}{1.6}
\centering
\resizebox{0.96\textwidth}{!}{
\begin{tabular}{c|c|c|c|c|c|c|c}
\hline
Index & \multicolumn{2}{c|}{Model}                                                                                                 & Parameter Number & Training Time & Training F1-Score & Testing F1-Score    & Inference Time (ms) \\ \hline
1     & \multicolumn{2}{c|}{AlexNet \cite{krizhevsky2012imagenet}}                                                                & 57.1M            & 37m 2s        & 95.53±0.21        & 94.48±0.53          & 188.5 ± 2.2         \\ \hline
2     & \multicolumn{2}{c|}{SqueezeNet \cite{iandola2016squeezenet}}                                                              & 743.1K           & 46m 7s        & 96.14±0.46        & 95.17±0.73          & 187.3 ± 1.6         \\ \hline
3     & \multicolumn{2}{c|}{GoogleNet \cite{szegedy2015going}}                                                                    & 5.6M             & 52m 28s       & 94.71±0.23        & 96.70±0.42          & 196.3 ± 0.5         \\ \hline
4     & \multicolumn{2}{c|}{Xception \cite{chollet2017xception}}                                                                  & 20.8M            & 89m 9s        & 94.87±0.29        & 96.16±0.55          & 211.3 ± 1.8         \\ \hline
5     & \multicolumn{2}{c|}{\textbf{DPN68} \cite{chen2017dual}}                                                                 & 11.8M            & 79m 10s       & 98.67±0.11        & \textbf{98.02±0.11} & 219.0 ± 6.9         \\ \hline
6     & \multicolumn{2}{c|}{MnasNet \cite{tan2019mnasnet}}                                                                        & 3.1M             & 51m 3s        & 91.95±0.30        & 94.78±1.06          & 191.2 ± 2.0         \\ \hline
7     & \multicolumn{1}{c|}{\multirow{3}{*}{ResNet}}       & \textbf{ResNet18} \cite{he2016deep}                                  & 11.2M            & 47m 30s       & 96.92±0.14        & \textbf{97.14±0.46} & 188.9 ± 0.9         \\ \cline{1-1} \cline{3-8} 
8     & \multicolumn{1}{c|}{}                              & \textbf{ResNet50} \cite{he2016deep}                                  & 23.5M            & 73m 17s       & 98.14±0.14        & \textbf{97.91±0.34} & 195.6 ± 0.4         \\ \cline{1-1} \cline{3-8} 
9     & \multicolumn{1}{c|}{}                              & \textbf{ResNet101} \cite{he2016deep}                                 & 42.5M            & 92m 55s       & 98.44±0.14        & \textbf{97.97±0.16} & 207.0 ± 0.6         \\ \hline
10    & \multicolumn{1}{c|}{\multirow{3}{*}{VGG}}          & VGG11  \cite{simonyan2014very}                                       & 128.8M           & 67m 46s       & 97.56±0.29        & 96.49±0.63          & 194.1 ± 1.3         \\ \cline{1-1} \cline{3-8} 
11    & \multicolumn{1}{c|}{}                              & VGG16  \cite{simonyan2014very}                                       & 134.3M           & 99m 25s       & 97.93±0.17        & 96.87±0.45          & 195.7 ± 1.4         \\ \cline{1-1} \cline{3-8} 
12    & \multicolumn{1}{c|}{}                              & \textbf{VGG19}  \cite{simonyan2014very}                              & 139.6M           & 112m 41s      & 97.91±0.11        & \textbf{97.01±0.57} & 197.2 ± 1.4         \\ \hline
13    & \multicolumn{1}{c|}{\multirow{3}{*}{Densenet}}     & \textbf{Densenet121} \cite{huang2017densely}                         & 7.0M             & 75m 40s       & 98.02±0.13        & \textbf{97.79±0.32} & 212.4 ± 0.8         \\ \cline{1-1} \cline{3-8} 
14    & \multicolumn{1}{c|}{}                              & \textbf{Densenet161 \cite{huang2017densely}}                         & 26.5M            & 133m 42s      & 98.48±0.16        & \textbf{97.85±0.27} & 227.4 ± 0.5         \\ \cline{1-1} \cline{3-8} 
15    & \multicolumn{1}{c|}{}                              & \textbf{Densenet169} \cite{huang2017densely}                         & 12.5M            & 85m 1s        & 98.24±0.16        & \textbf{97.85±0.27} & 226.8 ± 0.5         \\ \hline
16    & \multicolumn{1}{c|}{\multirow{3}{*}{Inception}}    & \textbf{Inception V3}  \cite{szegedy2016rethinking}                  & 24.4M            & 73m 50s       & 96.78±0.17        & \textbf{97.47±0.11} & 206.3 ± 0.4         \\ \cline{1-1} \cline{3-8} 
17    & \multicolumn{1}{c|}{}                              & Inception V4 \cite{szegedy2017inception}                             & 41.2M            & 120m 42s      & 95.88±0.20        & 96.93±0.15          & 235.4 ± 0.8         \\ \cline{1-1} \cline{3-8} 
18    & \multicolumn{1}{c|}{}                              & \multicolumn{1}{l|}{Inception-ResNet V2 \cite{szegedy2017inception}} & 54.3M            & 124m 36s      & 93.78±0.04        & 96.16±0.21          & 255.9 ± 1.4         \\ \hline
19    & \multicolumn{1}{c|}{\multirow{3}{*}{Mobilenet}}    & \textbf{Mobilenet V2} \cite{sandler2018mobilenetv2}                   & 2.2M             & 53m 27s       & 97.39±0.14        & \textbf{97.41±0.35} & 191.1 ± 0.8         \\ \cline{1-1} \cline{3-8} 
20    & \multicolumn{1}{c|}{}                              & MobilenetV3-small \cite{howard2019searching}                         & 1.5M             & 41m 27s       & 94.27±0.35        & 95.72±0.82          & 193.1 ± 1.2         \\ \cline{1-1} \cline{3-8} 
21    & \multicolumn{1}{c|}{}                              & MobilenetV3-large \cite{howard2019searching}                & 4.2M             & 49m 4s        & 96.69±0.19        & 96.89±0.54 & 193.8 ± 2.0         \\ \hline
22    & \multicolumn{1}{c|}{\multirow{6}{*}{EfficientNet}} & EfficientNet-b0 \cite{tan2019efficientnet}                           & 4.0M             & 63m 39s       & 93.10±0.32        & 95.95±0.91          & 202.0 ± 5.6         \\ \cline{1-1} \cline{3-8} 
23    & \multicolumn{1}{c|}{}                              & EfficientNet-b1 \cite{tan2019efficientnet}                           & 6.5M             & 77m 8s        & 93.74±0.18        & 95.80±0.70          & 203.8 ± 0.8         \\ \cline{1-1} \cline{3-8} 
24    & \multicolumn{1}{c|}{}                              & EfficientNet-b2 \cite{tan2019efficientnet}                           & 7.7M             & 78m 56s       & 94.16±0.22        & 96.36±0.58          & 204.5 ± 1.7         \\ \cline{1-1} \cline{3-8} 
25    & \multicolumn{1}{c|}{}                              & EfficientNet-b3 \cite{tan2019efficientnet}                           & 10.7M            & 92m 51s       & 95.11±0.33        & 96.74±0.47          & 211.3 ± 1.2         \\ \cline{1-1} \cline{3-8} 
26    & \multicolumn{1}{c|}{}                              & EfficientNet-b4 \cite{tan2019efficientnet}                           & 17.6M            & 113m 12s      & 94.23±0.69        & 95.74±0.69          & 216.3 ± 1.3         \\ \cline{1-1} \cline{3-8} 
27    & \multicolumn{1}{c|}{}                              & EfficientNet-b5 \cite{tan2019efficientnet}                           & 28.4M            & 144m 44s      & 94.07±0.27        & 95.57±0.53          & 224.1 ± 1.5         \\ \hline
\end{tabular}}
\label{tab:models}
\end{table*}